# Specialty detection in the context of telemedicine in a highly imbalanced multi-class distribution

Alaa Alomari[1,2*], Hossam Faris[1,2], Pedro A. Castillo[1],

**1** University of Granada, Granada, Spain
**2** Altibbi, King Hussain Business Park, Amman, 11831, Jordan

* omari@correo.ugr.es (AO)

## Abstract

The Covid-19 pandemic has led to an increase in the awareness of and demand for telemedicine services, resulting in a need for automating the process and relying on machine learning (ML) to reduce the operational load. This research proposes a specialty detection classifier based on a machine learning model to automate the process of detecting the correct specialty for each question and routing it to the correct doctor. The study focuses on handling multiclass and highly imbalanced datasets for Arabic medical questions, comparing some oversampling techniques, developing a Deep Neural Network (DNN) model for specialty detection, and exploring the hidden business areas that rely on specialty detection such as customizing and personalizing the consultation flow for different specialties. The proposed module is deployed in both synchronous and asynchronous medical consultations to provide more real-time classification, minimize the doctor effort in addressing the correct specialty, and give the system more flexibility in customizing the medical consultation flow. The evaluation and assessment are based on accuracy, precision, recall, and F1-score. The experimental results suggest that combining multiple techniques, such as SMOTE and reweighing with keyword identification, is necessary to achieve improved performance in detecting rare classes in imbalanced multiclass datasets. By using these techniques, specialty detection models can more accurately detect rare classes in real-world scenarios where imbalanced data is common.

## 1 Introduction

As Covid-19 invaded the world, speeding up the process of addressing the medical questions and providing healthcare support in a timely acceptable manner became a humanity target and public responsibility for governments and organizations. This forced the awareness about telemedicine and helped the telemedicine service providers like Teladoc, American Well (Amwell), Babylon Health, and Altibbi to have more popularity around the world.

Altibbi is a digital health cloud-based platform that focuses on telemedicine service for primary care purposes, Health Management Systems (HMS) and medical Arabic content, targeting the MENA region with more than 5 million structured medical consultation, more than 3 million accredited and verified medical content, and about 1 million electronic medical record (EMR). The telemedicine service is being provided through multiple channels like video, live chat, Global System for Mobile (GSM) calls, or asynchronous response.

To give more insights about Covid-19 impact on the awareness and need for telemedicine, Altibbi has made 2 million online medical consultations in 2020 compared to 1 million consultations from 2016 to 2019. This growth left no room for anything other than automating the process more and more and relying on Machine Learning (ML) in every possible way that could result in reducing the manual operational load. The exponential increase in the number of medical questions received by the patients that are being served in Altibbi on a daily basis has increased the urge to create a medical text classifier.

As primary care and day-to-day healthcare are the core of Altibbi telemedicine service, referring patients to the correct specialist is one of the most popular things that patients use Altibbi for. This comes mainly because most of the patients don't know exactly which specialist is best suited to their medical case.

This was easily diagnosed by the General Practitioner (GP) following the medical consultation with the patient. But as one of the medical services that Altibbi provides is asynchronously answered medical questions by specialized doctors within 24 hours, and as Altibbi is receiving thousands of questions like this to be answered asynchronously on a daily basis, and to avoid spamming the doctors' inboxes with questions that are not related to their specialities, Altibbi was doing the routing process manually by having some medical officers to review those questions and detect the correct speciality for each question and then to be routed to the correct doctor. This was a time-consuming and cost-inefficient process. In addition, the manual routing process was not real-time and not 100% accurate. This was because many questions were routed incorrectly due to the high intersection between keywords of the questions and the overlap of some specialties. This challenge highlights the need to automate the specialty detection stage by applying an automated detection system based on a machine learning model.

However, developing a machine learning based system for this task is not easy or straightforward.There are several challenges, including handling multiclass and highly imbalanced datasets. These are the core problems that will be addressed in this research.

This system is being proposed and deployed as an automated process in both synchronous and asynchronous medical consultations to help in:

- Providing a more real-time and accurate step for asynchronous questions.
- Minimizing the doctor's effort in addressing the correct speciality for primary care questions in the synchronous questions part.
- Giving Altibbi system more flexibility in customizing the medical consultation flow by following a specific decision tree of questions based on the detected specialty of the question body of the synchronous consultation.

In the course of this study, the various aspects of the specialty detection classifier will be examined, including its covert applications and potential to optimize the telemedicine process overall. This includes:

1. Handling skewed classes: classes with small or tiny distributions should not be neglected, as they represent patients' cases and health care.
2. Comparing some oversampling techniques.
3. Compare data-level oversampling with algorithmic-based oversampling.
4. Develop a DNN model for speciality detection, and apply some word embedding models.
5. Exploring the hidden areas that rely on specialty detection, such as:

(a) Customized and personalized consultation flow. For example, when a patient asks a question about obstetrics, the doctor should follow a specific clinical pathway (a multidisciplinary medical management tool used to manage the operational quality in healthcare concerning the standardization of the evaluation, diagnosis, and care of the patients with specific conditions) by asking a sequence of questions in a decision tree format such as date of the last period, number of normal births delivery versus number of caesarean sections, and others. For example, such a sequence of questions would not be appropriate for a pediatrician consultation. This research would help in exploring and automating personalizing the consultation flow for different specialities.

(b) From a commercial and business perspective, many pharmaceutical companies are showing a very high interest in sponsoring medical consultations that are concerning some specific specialties.

6. Evaluation and assessment: including accuracy, precision, recall, and F1-score.

The main contribution of this work can be summarized in two key aspects. Firstly, while most of researches in telemedicine and medical systems has primarily focused on the English language, a significant gap remains in addressing the unique challenges and requirements of Arabic medical content and consultations. To address this gap, this research leverages the extensive Altibbi Database, which is specifically designed for Arabic medical content and consultations. The inclusion of the Arabic language introduces distinct challenges, such as variations in dialects and a scarcity of research tailored specifically to Arabic medical content. By undertaking this study, we aim to bridge this gap and contribute to the expanding knowledge in telemedicine by providing insights into the development of a classification system within the Arabic medical context. Furthermore, an additional noteworthy contribution of this work lies in addressing the crucial issue of imbalanced classification for rare specialties in the medical field. Dealing with imbalanced data is particularly significant in healthcare, where certain specialties may have limited samples or face a scarcity of research compared to others. By employing appropriate techniques, such as oversampling, reweighing, and keyword identification, this research aim to tackle the challenge of imbalanced classification and enhance the accuracy and effectiveness of specialty detection within the Arabic telemedicine framework.

The rest of the paper is organized as follows: Next section, background and related works, provides a comprehensive overview of the challenges, techniques and related works. Then, in section 3, the methodology and the proposed approach of the model will be presented. Section 4 presents the experimental setup and obtained results. Finally, the conclusions and future work are discussed in Section 5.

## 2 Background and related works

In the field of machine learning, there are several challenging problems that many researchers have studied in detail. These include the handling of highly imbalanced datasets, as well as the analysis of highly multiclass datasets. Additionally, some researchers have focused on developing machine learning models tailored to the Arabic language and its many dialects, which present unique challenges due to their diversity. These are important areas of research that have the potential to improve the accuracy and applicability of machine learning models in a wide range of real-world scenarios.

## 2.1 Approaching highly multiclass datasets

A smaller number of options generally leads to a more accurate classification task [6]. Classification can be either binary or multinomial (multiclass). In this research, the classification problem is multiclass, with a high likelihood of overlapping and intersecting specialties. However, at the end, the telemedicine doctor will give you a recommendation to visit a doctor with specific speciality if needed. Therefore, regardless of the overlapping and possibility of having multi-labels for a given case, the classification model has to deal with it as multiclasses not a multi-labels problem based on the assumption that each medical question will be routed to only one specialist.

## 2.2 Handling highly imbalanced-multiclass dataset

Similarly, human beings' final decisions or votes for the best option are not necessarily the outcome of their conscious decisions. Their decisions could be influenced by many factors and subtle sway around them, such as unconscious thinking processes, emotions, first impressions, preconceptions, or following majority voting [7, 8]. Just as the imbalance of observations in the real world could affect the trend of taking decisions for human beings, machine learning algorithms are mimicking the same experience and would be affected by imbalanced data sets. Based on International Data Corporation (IDC), a market research company, about 1.2 zettabytes (1.2 trillion gigabytes) of new data were created in 2010. This amount of new data was predicted to grow exponentially generating 175 trillion gigabytes of new data around the world in 2025 [9]. The tremendous growth in data generation around the world and mainly in the technical field is increasing the gap of imbalanced data sets and giving more urge to handle this issue in a more strategic and programmatic way.

Kaur et al. in [21] presented a detailed survey that highlighted the different factors that contribute to imbalanced data, including data collection bias, and how it can negatively affect the performance of machine learning models by providing a poor accuracy carried out on minority class. They also provided a general summary of the various methods utilized to address imbalanced datasets, such as pre-processing methods like resampling techniques, algorithmic centered approaches like cost-sensitive learning and One-Class Learning, and hybrid approaches to minimize chances of information loss and to provide prediction. Moreover, they dived deeply into the applications of imbalanced data across various domains like healthcare, fraud detection, and social media analysis and carried out an in-depth analysis of the challenges and opportunities in each of these domains.

Yu in [22] presented an experimental results on four high-dimensional and imbalanced biomedical datasets. The datasets used in the experiments are Colon, Lung, Ovarian I, and Ovarian II, which were collected from colon cancer patients, non-small cell lung cancer patients, and women with ovarian cancer. The evaluation criteria used in the experiments include overall accuracy, true positive rate, true negative rate, F-measure, G-mean, and area under the receiver operating characteristic curve (AUC). In this research AUC has been recognized to be considered a reliable performance measure for class imbalance problems. However, it is worth mentioning that ROC curves are frequently employed to illustrate outcomes in binary decision tasks within machine learning. Nonetheless, when handling imbalanced datasets, Precision-Recall (PR) curves offer a more informative depiction of an algorithm's effectiveness [30]. In addition, the AUC metric needs to be modified in order to be applied for multi-class problems. This explains why it is less common to be adopted in such cases.

Yu in [22] proposed a novel and hybrid ensemble learning solution called asBagging_FSS (asymmetric bagging ensemble classifier with feature subspace (FSS)). In this method they utilized clustering to filter redundant and feature selection to filter noisy features. Finally they compared its performance with eight other classification approaches. Gaussian radial basis function-based SVM was used as the base classifier.

However, many other previous works were conducted to address the imbalanced datasets either from data level perspective or from some algorithms perspective [10]. As for managing imbalanced dataset on the data level, different forms of resampling have been used to manage imbalanced datasets, such as:

1. Random oversampling [11]: which is to randomly select entries from minority classes and replicate them and add them to the training dataset. This method is considered as a naive resampling method as it has no preferences and zero assumption about data points being duplicated and no heuristics are used.

2. SMOTE (Synthetic Minority Oversampling Technique) [3, 11]: it is a method originally introduced by Chawla et al. [11], aimed at mitigating the challenges posed by imbalanced datasets. It involves generating synthetic instances by interpolating between existing minority class samples based on the K-nearest neighbors (KNN) algorithm. This approach has been widely adopted by researchers to address practical problems. For instance, Akbar et al. [25] proposed an innovative strategy called iAFP-gap-SMOTE, which integrates feature extraction and oversampling techniques to enhance the identification of Antifreeze proteins (AFPs). AFPs play a crucial role in enabling organisms to survive in extreme cold environments. By combining the strengths of SMOTE with iAFP-gap-SMOTE, researchers achieved improved performance in accurately identifying AFPs, thus contributing to the advancement of understanding these vital proteins and their functionalities in extreme temperature conditions.

3. Random undersampling [11]: this mainly can be done by randomly selecting examples (entries) from the majority class of the training set and deleting them. This method could lead to the loss of representative information from the training set. As in the Random Oversampling case, the random undersampling paradigm follows the same concept of no heuristics being used, which leads to consider Random Undersampling in the same category of naive resampling process.

4. Direct Undersampling [12, 13]: in which the items to be eliminated from the dataset are informed. In other words, this method classifies the items into borderline, noise or far from the decision border. Thus, such undersampling methods drop the noise or far items as they mostly have less impact and importance on the dataset [14].

5. Oversampling with informed generation of new samples [12, 13]: where the new items to be added to the dataset are coming from informed decisions.

6. Mix of minority oversampling and majority undersampling. [11]

On the other hand, many other researches were presented to handle this problem from algorithmic perspective like:

1. Cost adjustment of the classes [13, 15]: genetic programming can be used to assign different costs to different types of misclassification errors. This can lead to improved precision (by increasing the penalty on false positives) or improved recall (by increasing the penalty on false negatives).

2. Probabilistic adjustment [13, 15, 16]: it is mainly used when working with decision trees by adjusting the probabilistic estimate at the tree leaf. There are two main techniques that can be used to implement this probabilistic adjustment which are smoothing and curtailment.

As for Smoothing, it can be implemented either by running Laplace Correction Method which mainly works when the problem is for two classes and it tries to make the probability something around 0.5 [17]. The other smoothing method is called m-estimation which is an unconditional probability method that use the formula of P'= (k+b*m)/(n+m) as the probability estimate, where b is base rate of the positive cases and m is the shift controller parameter that controls how much scores are shifted towards b and it is usually chosen using cross validation [16]. On the other hand, curtailment method works by eliminating some leafs and keep others based on the number of training examples associated with each child in the decision tree [16].

3. Decision threshold adjustment [13, 15]: in this boolean function, a threshold value will be adjusted to specify whether a neuron will be activated or not.

4. Recognition based learning virsus discrimination based learning [13, 15]: which is mainly learning from one class rather than discrimination-based learning.

Practically, most of the imbalanced dataset problems are about binary classification [18] like fraud detection, spam filtration, benign versus malignant, etc. However, this research will cover and address how to deal with imbalanced datasets for multiclass classification problems. In this research, some of the previously mentioned approaches will be explored to validate the applicability of them in deep learning for handling the highly imbalanced multiclass classification problem.

## 2.3 Arabic language and the variance of its dialects

One of the challenges that arises when building a machine learning module for the Arabic language is handling the different dialects of Arabic. The complexity of the task increases due to the unique features and nuances of the language, such as the lack of standardization in spelling and grammar and the wide range of dialects. In this research, doctors and patients come from a variety of backgrounds and speak different dialects, including those from Saudi Arabia, Jordan, Egypt, Iraq, Libya, and others.

Hammoud et al. [19] focused on Named Entity Recognition (NER), and information extraction in the context of Arabic medical text. Their approach was to use various machine learning techniques, such as feature engineering and a conditional random field (CRF) model. They also described the development of a corpus of labeled data, which was used to train and evaluate their models. They reported high precision and recall values for entity recognition, as well as the successful extraction of various types of medical information from the text.

Other researchers, such as Alanazi in [20], have followed a hybrid approach by combining rule-based and machine learning techniques, such as Bayesian Belief Networks (BBN) to recognize named entities in Arabic medical text with some dependency on another factor of relying on hand-made linguistic rules.

## 2.4 Comparative analysis

In the field of text classification, previous research has explored various approaches to tackle challenges related to the classification of specialized domains, such as Arabic medical text. Al-Radaideh et al. [26] proposed an associative rule-based classifier specifically designed for Arabic medical text classification. Their approach leveraged the inherent associations between medical concepts in the text to make accurate classifications. While their work focused on the classification of medical text using association rules, this research addresses the problem of specialty detection on imbalanced multi-class datasets.

In contrast to Al-Radaideh's approach [26], which employed associative rules, this study focuses on machine learning techniques, specifically utilizing the BILSTM model. This research examines the effectiveness of various techniques, including reweighing, oversampling, and keyword identification, to improve the performance of specialty detection models on imbalanced multi-class datasets.

Furthermore, in the evaluation of oversampling techniques phase, this research compared the performance of SMOTE and ADASYN, two commonly used approaches for addressing class imbalance. The experimental results demonstrated that SMOTE outperformed ADASYN in our specific domain of specialty detection.

Additionally, this study introduced the concept of reweighing the rare classes based on the presence of a keyword, which proved to be effective in our experiments. This technique yielded better results than SMOTE, indicating the importance of considering keyword identification for improving performance on imbalanced datasets.

While Al-Radaideh et al. [26] study primarily focused on associative rule-based classification for Arabic medical text, this research contributes to the field by addressing the challenges of specialty detection on imbalanced multi-class datasets using machine learning techniques and a combination of reweighing, oversampling, and keyword identification.

Overall, this comparative analysis highlights the different focuses and approaches between Al-Radaideh's work [26] and our research, providing insights into the unique contributions and relevance of our study in the context of specialty detection in imbalanced multi-class datasets.

# 3 Methodology

The proposed approach addresses three main challenges: handling multiclass, dataset imbalance, and dialect processing. Fig 1 shows a detailed flow of the proposed approach to address these challenges.

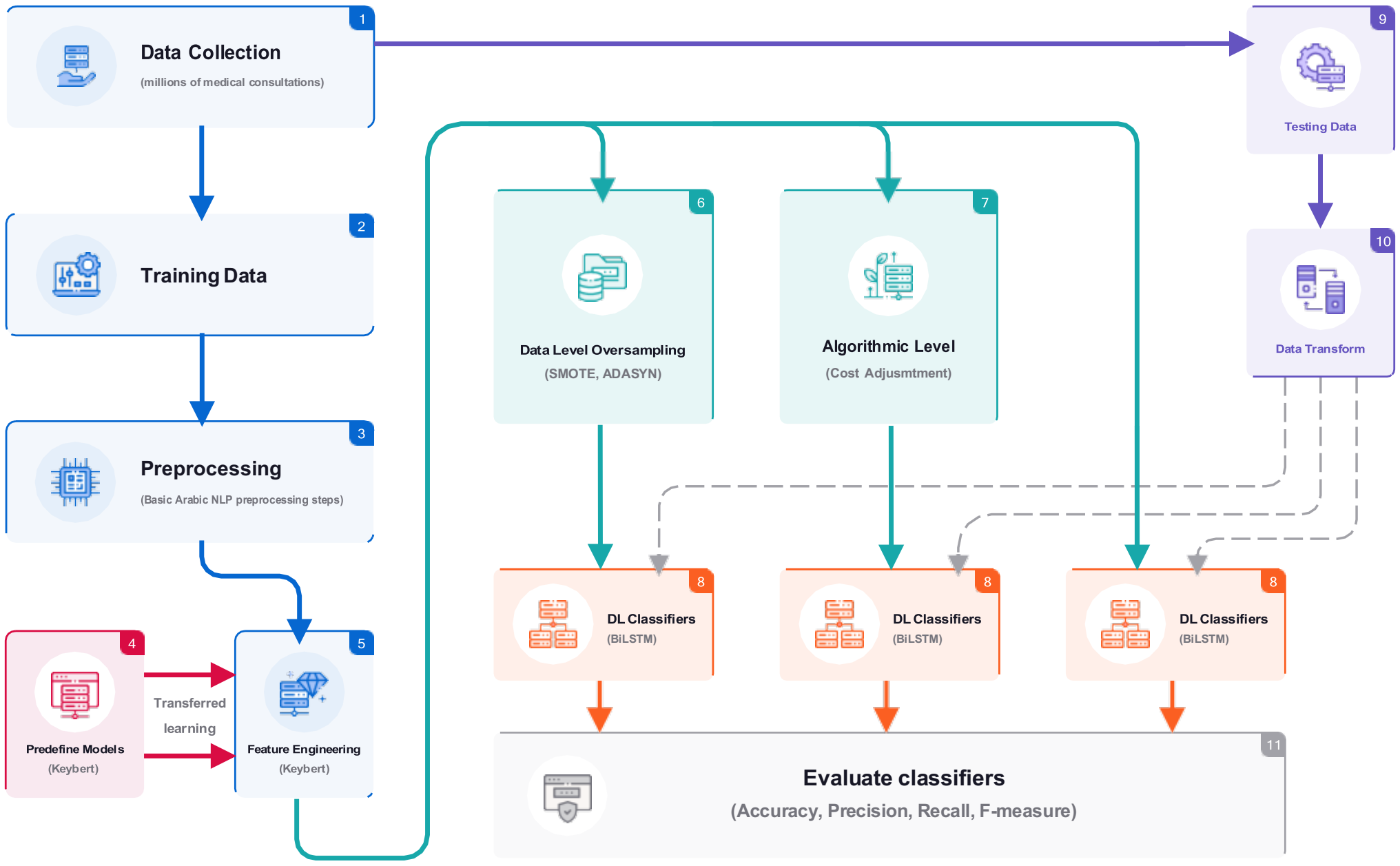

**Fig 1. Schematic diagram depicting the research methodology.**

### 3.1 Data collection

Fig 2 shows the imbalanced distribution of medical consultations in Altibbi database from September 2016 to November 2022. There were a total of 526K consultations, which represent about 10% of the total number of consultations on Altibbi. The telemedicine doctor recommends the patient to refer to a specialist if their medical consultation was not resolved fully through the virtual consultation. It is worth mentioning here that this research involves the use of data collected from users of Altibbi Platform. The consent was obtained from those users through the Terms and Conditions agreement on the platform, in which users authorize Altibbi to use their data in the developed machine learning models for the purpose of research and improving the quality of health care service. Taking in consideration that all user data was anonymized and de-identified to protect the privacy of Altibbi users. Also, this research followed all relevant ethical guidelines for conducting research with human subjects, including obtaining informed consent, ensuring confidentiality of participant data, and minimizing the risk of harm to participants.

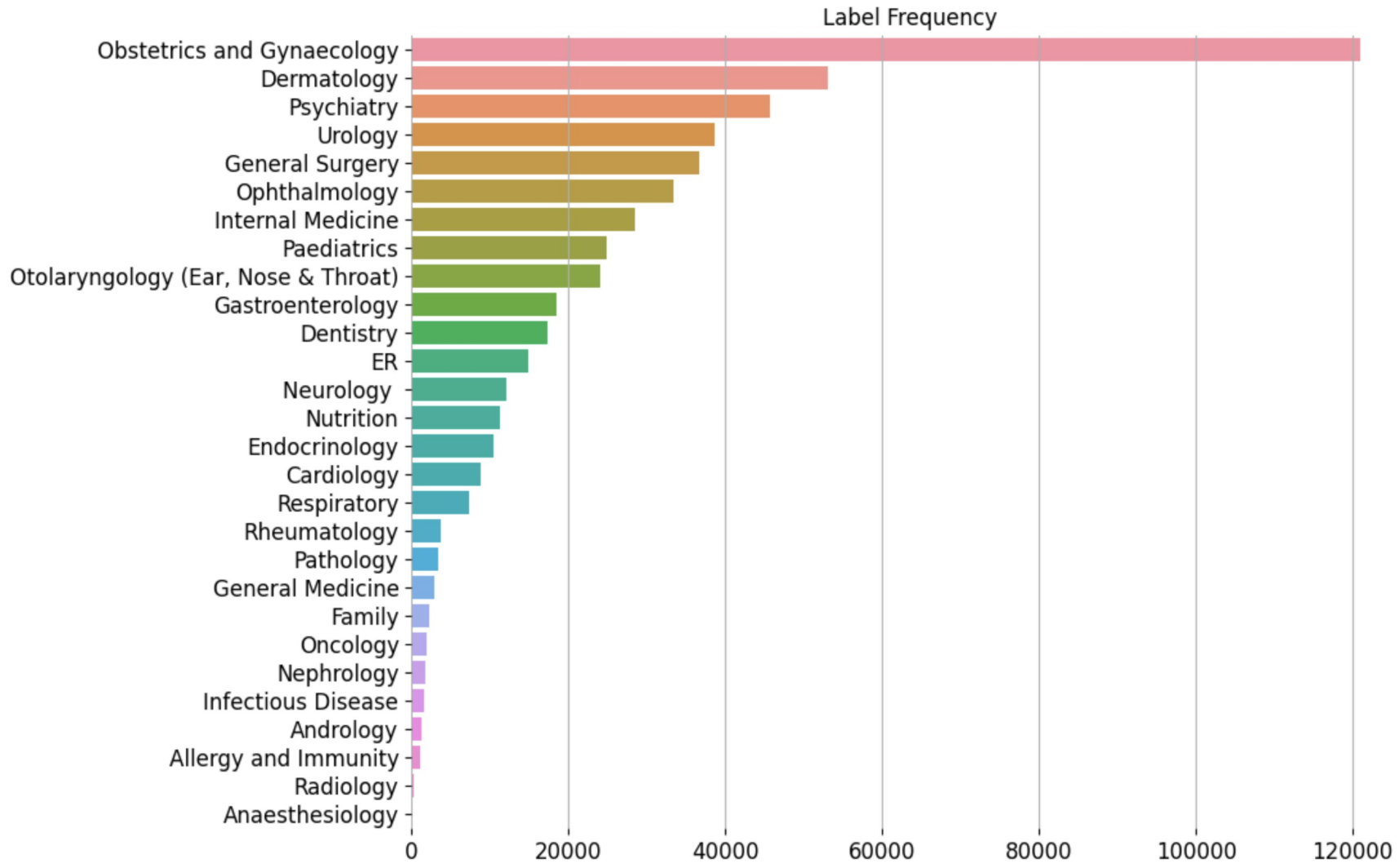

**Fig 2. Distribution of the free questions asked on Altibbi platform over their medical specialties.**

### 3.2 Preprocessing

This stage is to prune and clean the training set by removing stop words, removing diacritics, and other non-alphabetical characters, as well as to apply stemming as needed. Removing duplicate content will utilize Bag-of-Words (BoW) and TF-IDF. BoW is simply a list of vectors containing the count of words in the document while TF-IDF is a more sophisticated model that takes into account the frequency of a word in a document and the frequency of the word in the corpus as a whole. This allows TF-IDF to identify words that are important in a document, even if they do not appear very often. TfidfVectorizer from sklearn will be used for feature extraction for this regard.

### 3.3 Predefined models

This research will utilize and use some libraries like KeyBert and open source pre-trained distributed word embedding like AraVec v3 and AraBERT v2. KeyBERT is a Python library for keyword extraction using BERT (Bidirectional Encoder Representations from Transformers) embedding. It provides a simple interface for extracting keywords or keyphrases from the consultation/question body using unsupervised machine learning techniques. KeyBERT utilizes BERT to generate embeddings for each word or token in the medical question body, and then calculates the similarity between each word and the entire question to determine the most relevant words or phrases. AraVec has been built using Twitter tweets and Wikipedia Arabic articles as input source for its training set, while AraBERT dataset sources were OSCAR unshuffled and filtered, Arabic Wikipedia, The 1.5B words Arabic Corpus, The OSIAN Corpus, and Assafir news articles. Such pre-trained embedding models help in representing the words as vectors in a continuous space which helps in building the semantic and syntactic relations between words. Fine tuning will be applied on those models to transfer the learning into a more specialized model.

## 3.4 Feature engineering

As BoW and TF-IDF in the preprocessing step don't preserve the relationship between words and can't capture the meaning of consecutive words; Word2Vec model will be used as a word embedding model which helps in capturing whether words appear in similar contexts (inter-word semantic) as well as to reduce dimensionality. By taking the fact that Word2Vec is a predictive model, the research will make a comparison with and shed some light on the GloVe embedding model which relies on count-based instead of prediction to find the co-occurrence of the words. In the end, the BERT from Google will be used in this research. The BERT from *ktrain library* (which is a lightweight wrapper for the deep learning library TensorFlow Keras) to show its power in semantic analysis by taking into consideration the context for each occurrence of a given word.

## 3.5 Oversampling

This research will use data and algorithmic level oversampling techniques. The effectiveness of both SMOTE and Tomek Links and the combination of them will be examined to manage the imbalance of data distribution for the dataset that has been obtained from Altibbi. Other oversampling techniques, such as ADASYN, will also be tested for comparison purposes.

On the other hand, as for algorithmic level oversampling, cost adjustment of the classes will be applied to compare their performance in manipulating data imbalancement with data level oversampling techniques.

## 3.6 Deep learning classifier

As understanding the question body and classifying it relies heavily on its semantic and harmony in its words and phrases and thus understanding each word relies on its previous word(s), Recurrent Neural Network (RNN) would be the best fit for such a problem. Thus, LSTM and BiLSTM will be used from Keras side by side for applying some other models like *Sequential* to allow creating neural network objects with a sequence of layers, and *Dropout* to manage overfitting and regularization.

## 3.7 Data transform

While many scalars like StandardScaler, MinMaxScaler, MaxAbsScaler, RobustScaler, PowerTransformer, QuantileTransformer with uniform output, QuantileTransformer with Gaussian output, and Normalizer could be used, MinMax Scalar will be used as it helps in maintaining the original distribution of the dataset and doesn't reduce the importance of the outliers and anomalies. Fit mainly means getting the minimum and maximum value in the training set, and *Transform* means applying the formula of *Xi - min(x) / (max(x) - min(x))* It is very clear that only Transform will be applied on the *Testing set* to follow the same scalar of the training set where Fit transform has been applied there.

### 3.8 Evaluate classifiers

To test the effectiveness of the proposed approach, the following criteria have been taken in consideration:

1. K-fold cross-validation (CV): this approach splits the original dataset into K equal-sized subsets or "folds". The model is then trained on K-1 of these folds and validated on the remaining fold. This process is repeated K times, with each fold serving as the validation set once. An 80/20 stratified sampling ratio will be applied to guarantee testing all classes in the test set.

2. Confusion matrix, which is a table that summarizes the performance of a classification model by comparing predicted class labels with the actual class labels of a test dataset. The confusion matrix is often used to calculate accuracy, recall, and precision by applying formulas on true positives (TP), false positives (FP), true negative (TN), and false negative (FN).

   (a) Accuracy: is the proportion of TP and TN to the total number of instances in the dataset.

   $$\text{Accuracy} = \frac{TP + TN}{TP + FP + TN + FN} \quad (1)$$

   (b) Recall: is the true positive rate. It is the proportion of true positives to the total number of actual positives.

   $$\text{Recall} = \frac{TP}{TP + FN} \quad (2)$$

   (c) Precision: is the positive predictive value. It is the proportion of TP to the total number of positive predictions.

   $$\text{Precision} = \frac{TP}{TP + FP} \quad (3)$$

3. F1-Score: this metric combines the precision and recall metrics into a single score ranges from 0 to 1, with a score of 1 indicating perfect precision and recall. A higher F1 score indicates a better overall performance.

$$F1 = \frac{2 \times Precision \times Recall}{Precision + Recall} \quad (4)$$

## 4 Experimental results

In order to develop the experimental procedure for classifying Arabic healthcare questions, this research drew upon the findings from a previous research on classification of Arabic healthcare questions based on word embedding learned from massive consultations. Specifically, BILSTM model has been chosen exclusively, as the previous experiments with LSTM produced unsatisfactory results [24]. The procedure in this study involved several key phases, including data preprocessing, model development, and evaluation. To prevent overfitting, optimize the model's performance, and mitigate the impact of local minima during training, early stopping was implemented. A dataset of over 523,000 consultations from the Altibbi Telemedicine Database has been used, encompassing 45 different medical specialties. In evaluating the models, a variety of metrics, including precision, recall, F1-score, and accuracy have been taken in consideration.

### 4.1 Experiments setup

In terms of hyperparameter selection, a fine-tuning approach was employed during the training of our deep learning-based model. The learning rate hyperparameter was explored by considering various values such as 0.05, 0.1, 0.2, and 0.5, with reasonable intervals between these rates. The number of epochs, denoting the complete traversal of the entire training dataset, was adjusted in conjunction with the early stopping hyperparameter to mitigate the risks of underfitting or overfitting. Additionally, the number of nodes in the model was experimented with, encompassing configurations of 15, 30, 64, and 128 neurons. Through this iterative process of hyperparameter tuning, we aimed to optimize the performance and generalization capabilities of our model.

The experiment was conducted using Python 3.10.12 as the programming language and several essential Python libraries. The pandas library (version 1.4.4) was utilized for data manipulation and preprocessing tasks, facilitating efficient handling of the datasets. For implementing and training the deep learning-based model, we employed tensorflow (version 2.11.0), a powerful framework known for its extensive support of neural network architectures. The numpy library (version 1.24.2) was instrumental in performing numerical computations and array operations, ensuring smooth processing of the model's computations. By leveraging these Python packages, we ensured a robust and well-supported environment for our experimental setup, enabling seamless implementation and evaluation of our deep learning model.

### 4.2 Basic model implementation without handling imbalance

In the first phase of the experimentation, the AltibbiVec model [23] was used to build and test BILSTM on the imbalanced dataset. The dataset was split into training and testing sets in an 80/20 stratified sampling ratio as shown in the pie chart in Fig 3.

In this phase BILSTM model with 15, 30, 64, and 128 nodes with minimal fine tuning. The best result was for BILSTM with 128 units, where the precision was 0.421, recall was 0.356, f1-score was 0.358 and accuracy was 0.652. The final result of the comparison is as shown in Table 1:

**Table 1.** Build and experiment different BILSTM nodes on the imbalanced dataset with fine tuning.

| Model | Precision | Recall | F1-score | Accuracy |
|---|---|---|---|---|
| BILSTM 15 | 0.39 | 0.33 | 0.33 | 0.64 |
| BiLSTM 30 | 0.39 | 0.33 | 0.33 | 0.647 |
| BiLSTM 64 | 0.409 | 0.343 | 0.343 | 0.651 |
| BILSTM 128 | **0.421** | **0.356** | **0.358** | **0.652** |

### 4.3 BILSTM with oversampling techniques

During the oversampling comparison phase, the effectiveness of BILSTM neural networks with different numbers of units (15, 30, 64, and 128 units) and two oversampling techniques (SMOTE and ADASYN) were evaluated for a multi-class

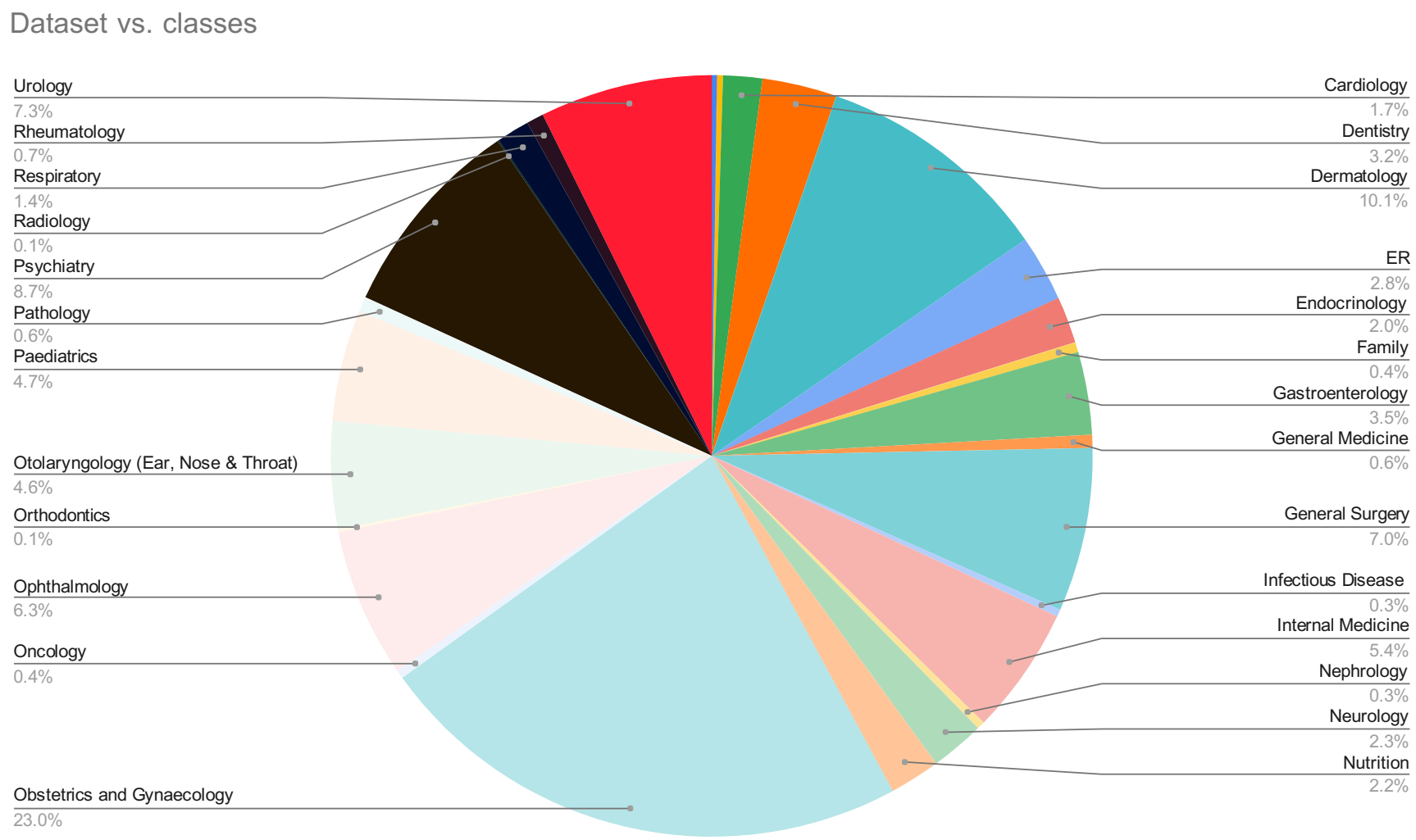

**Fig 3. Classes distribution in the dataset.**

imbalanced dataset. The results showed that the use of SMOTE led to an improvement in recall, indicating its ability to effectively address the issue of class imbalance. The findings are presented in Table 2:

**Table 2.** BILSTM using SMOTE and ADASYN oversampling techniques.

| Model | Precision | Recall | F1-score |
|---|---|---|---|
| BILSTM 15 SMOTE | 0.222 | 0.259 | 0.229 |
| BILSTM 15 ADASYN | 0.272 | 0.279 | 0.271 |
| BILSTM 30 SMOTE | 0.258 | **0.299** | 0.265 |
| BILSTM 30 ADASYN | 0.213 | 0.236 | 0.213 |
| BILSTM 64 SMOTE | 0.232 | 0.245 | 0.224 |
| BILSTM 64 ADASYN | 0.238 | 0.271 | 0.238 |
| BILSTM 128 SMOTE | 0.250 | 0.277 | 0.251 |
| BILSTM 128 ADASYN | 0.232 | 0.259 | 0.233 |

## 4.4 BILSTM with weight adjustment for rare classes

Furthermore, to exploring the effectiveness of different oversampling techniques and neural network architectures in addressing multi-class imbalanced datasets, this study also investigated the impact of assigning higher weights to rare classes. Rare classes were defined as classes with less than 1000 instances. The BILSTM model with 15 units and weighted training demonstrated the best recall performance in this scenario. These findings suggest that combining SMOTE and weighted training may be a promising approach to enhance the performance of BILSTM models when dealing with imbalanced multi-class datasets, particularly those with rare classes. The results of weight adjustment for rare classes are presented in Table 3.

## 4.5 Specialty keyword identification

Finally, specialty keyword identification phase has taken place by fine tuning Bert language model using unsupervised learning on Altibbi content which resulted in

**Table 3.** Build and experiment BILSTM models with the addition of eeightings on the classes.

| Model | Precision | Recall | F1-score | Accuracy |
|---|---|---|---|---|
| BILSTM 15 | 0.319 | 0.397 | 0.294 | 0.467 |
| BiLSTM 30 | 0.326 | 0.411 | 0.320 | 0.521 |
| BiLSTM 64 | 0.334 | 0.410 | 0.318 | 0.488 |
| BILSTM 128 | 0.335 | 0.409 | 0.330 | 0.536 |

AltibbiBert. Then, AltibbiBert was embedded in KeyBert in order to do the keywords extraction that are highly related to the rare specialties. A list of the main keywords was created for each of the rare specialties and got validated by subject matter expert with two expert doctors from Altibbi. Table 4 shows a sample of the keywords that were extracted and correlated to the rare classes.

**Table 4.** Sample of specialty keyword identification for rare specialties.

| Speciality/Keyword | Keyword 1 | Keyword 2 | Keyword 3 | Keyword 4 | Keyword 5 |
|---|---|---|---|---|---|
| Anaesthesiology | بنج | التخدير | الاوكسجين | موضعي | جراحيه |
| | Anesthetic | Anesthesia | Oxygen | Local | Surgical |
| Allergy and Immunity | حساسيه | حكه شديده | احمرار | ضيق | الربو |
| | Allergy | Intense Itching | Redness | Shortness | Asthma |
| Infectious Disease | انتفاخ | ميكوبلازما | جرثومه | حلزونيه | المضادات |
| | Swelling | Mycoplasma | Microbe | Spiral | Antibiotics |
| Nephrology | الكلي | المثانه | حرقه البول | التهاب بروستات | الكرياتين |
| | Kidneys | Bladder | Burning Urine | Prostatitis | Creatine |
| Oncology | اورام | كتله | الكيماوي | خبيث | الماموقرام |
| | Tumors | Mass | Chemical | Malignant | Mammograms |
| Pathology | الصفائح | الهوميغلبين | اوعيه دمويه | الغدد | كرات الدم |
| | Plates | Homeoglibin | Blood Vessels | Glands | Blood Cells |
| Rheumatology | الركبه | المفاصل | دسك | الفقري | الرقبه |
| | Knee | Joints | Prolapsed Disc | Vertebral | Nick |

As for the result of running the model after keyword identification phase, Table 5 shows the result for BILSTM with 15 nodes. Different reweighing factors of 2, 5, 10, and 15 are applied to the positive class (keyword-present) in the dataset during training to determine the optimal weight for improving the performance of the machine learning model in identifying the positive class (i.e., the keyword-present samples) which compensates for the fact that the positive class may have fewer samples than the negative class. Below was the result for BILSTM with 15 nodes for the above mentioned factors.

**Table 5.** Keyword identification with different factors.

| Factor | Precision | Recall | F1-score | Accuracy |
|---|---|---|---|---|
| 2 | 0.374 | 0.343 | 0.338 | 0.638 |
| 5 | 0.386 | 0.339 | 0.337 | 0.635 |
| 10 | 0.365 | 0.338 | 0.335 | 0.638 |
| 15 | 0.390 | 0.338 | 0.332 | 0.636 |

To facilitate the comparison process, precision, recall, and f1-score for rare classes - those with a population of less than 1000 instances - with different reweighting factors are presented in Table 6

## 4.6 Experimental summary

A summary of the conducted experiments on addressing imbalanced data sets in Arabic medical consultations is presented in Table 7. The findings reveal that the optimal

**Table 6.** Results of BILSTM for rare classes for the imbalanced dataset with different reweighting factors.

| BILSTM 15 | Precision | Recall | F1-score |
|---|---|---|---|
| Imbalanced | 0.143 | 0.015 | 0.030 |
| Factor 2 | 0.200 | 0.025 | 0.044 |
| Factor 5 | 0.197 | 0.019 | 0.033 |
| Factor 10 | 0.104 | 0.008 | 0.015 |
| Factor 15 | 0.256 | 0.025 | 0.039 |

outcome was achieved with BILSTM 15 with reweighing factor of 15 for keywords of the rare classes.

**Table 7.** Summary table: oversampling techniques using SMOTE, ADASYN, and weighted rare classes.

| **Model** | Precision | Recall | F1-score |
|---|---|---|---|
| BILSTM 15 SMOTE | 0.222 | 0.259 | 0.229 |
| BILSTM 15 ADASYN | 0.272 | 0.279 | 0.271 |
| BILSTM 15 Weighted | 0.045 | 0.321 | 0.074 |
| **BILSTM 15 Factor 15** | **0.390** | **0.338** | **0.332** |
| BILSTM 15 imbalanced | 0.155 | 0.008 | 0.016 |
| BILSTM 30 SMOTE | 0.258 | 0.299 | 0.265 |
| BILSTM 30 ADASYN | 0.213 | 0.236 | 0.213 |
| BILSTM 30 Weighted | 0.053 | 0.294 | 0.085 |
| BILSTM 30 imbalanced | 0.171 | 0.028 | 0.045 |
| BILSTM 64 SMOTE | 0.232 | 0.245 | 0.224 |
| BILSTM 64 ADASYN | 0.238 | 0.271 | 0.238 |
| BILSTM 64 Weighted | 0.048 | 0.308 | 0.08 |
| BILSTM 64 imbalanced | 0.187 | 0.019 | 0.032 |
| BILSTM 128 SMOTE | 0.250 | 0.277 | 0.251 |
| BILSTM 128 ADASYN | 0.232 | 0.259 | 0.233 |
| BILSTM 128 Weighted | 0.052 | 0.267 | 0.084 |
| BILSTM 128 imbalanced | 0.217 | 0.045 | 0.065 |

## 5 Conclusion and future works

The experimental results indicate that several techniques are necessary to improve the performance of a specialty detection machine learning model on imbalanced multi-class datasets. These techniques include reweighing, oversampling, and keyword identification. The specific choice of BILSTM units (15, 30, 64, and 128) did not have a significant impact on the performance of the model.

In terms of oversampling, SMOTE and ADASYN were both used to address class imbalance, but SMOTE performed better than ADASYN. The results also showed that reweighing the rare classes based on the presence of a keyword gave better results than SMOTE. Finally, factoring keyword-present with a factor of 15 gave the best result among the different factors tested (2, 5, 10, and 15).

Overall, these findings suggest that combining multiple techniques for addressing class imbalance is necessary to achieve improved performance in the detection of rare classes in imbalanced multi-class datasets. The most effective techniques identified in this study were SMOTE and reweighing with keyword identification. By using these techniques, specialty detection models can more accurately detect rare classes in real-world scenarios where imbalanced data is common.

In terms of future work, it is recommended to explore the use of additional deep learning models, such as Convolutional Neural Networks (CNNs) and Transformer-based models, as well as alternative methods for keyword identification, such as Named Entity Recognition (NER) or other pre-trained language models like UniLM, T5 (Text-to-Text Transfer Transformer), or GPT. By investigating these options, it may be possible to further improve the performance of the specialty detection model and expand the range of applications in real-world scenarios where imbalanced data is common. Furthermore, in the realm of future work, incorporating predictor algorithms such as Deep-AntiFP [27] and iHBP-DeepPSSM [28] can be explored to enhance the specialty detection model's performance. These predictor algorithms leverage advanced techniques and methodologies, such as deep learning architectures and protein sequence-based feature extraction, to accurately predict protein functionalities and attributes. By integrating these predictor algorithms into the existing framework, a more comprehensive and robust specialty detection model can be developed. This expansion in methodology could lead to improved accuracy and broader applicability in real-world scenarios characterized by imbalanced data.

## Acknowledgements

This work was supported by the Ministerio Español de Ciencia e Innovación under project number PID2020-115570GB-C22 MCIN/AEI/10.13039/501100011033 and
by the Cátedra de Empresa Tecnología para las Personas (UGR-Fujitsu).